\documentclass{article}

\PassOptionsToPackage{numbers, compress}{natbib}
\usepackage[final]{neurips_data_2023}

\usepackage[utf8]{inputenc} % allow utf-8 input
\usepackage[T1]{fontenc}    % use 8-bit T1 fonts
\usepackage{hyperref}       % hyperlinks
\usepackage{url}            % simple URL typesetting
\usepackage{booktabs}       % professional-quality tables
\usepackage{amsfonts}       % blackboard math symbols
\usepackage{nicefrac}       % compact symbols for 1/2, etc.
\usepackage{microtype}      % microtypography
\usepackage{xcolor}         % colors
\usepackage{graphicx}
\usepackage{caption}
\usepackage{subcaption}
\usepackage{hyperref}
\usepackage[numbers]{natbib} 
\usepackage{authblk}
 
\hypersetup{
    unicode=false,          % non-Latin characters in Acrobat’s bookmarks
    pdftoolbar=true,        % show Acrobat’s toolbar?
    pdfmenubar=true,        % show Acrobat’s menu?
    pdffitwindow=false,     % window fit to page when opened
    pdfstartview={FitH},    % fits the width of the page to the window
    pdftitle={},    % title
    pdfauthor={},     % author
    pdfsubject={},   % subject of the document
    pdfcreator={},   % creator of the document
    pdfproducer={}, % producer of the document
    pdfkeywords={keyword1} {key2} {key3}, % list of keywords
    pdfnewwindow=true,      % links in new window
    colorlinks=true,       % false: boxed links; true: colored links
    linkcolor=blue,          % color of internal links (change box color with linkbordercolor)
    citecolor=blue,        % color of links to bibliography
    filecolor=blue,      % color of file links
    urlcolor=blue           % color of external links
}

\title{Mesogeos: A multi-purpose dataset for data-driven wildfire modeling in the Mediterranean}

\author[1, 2]{Spyros Kondylatos}
\author[1, 2]{Ioannis Prapas}
\author[2]{Gustau Camps-Valls}
\author[1]{Ioannis Papoutsis}

\affil[1]{Orion Lab, Institute for Astronomy, Astrophysics, Space Applications, and Remote Sensing, National Observatory of Athens}
\affil[2]{Image Processing Laboratory (IPL), Universitat de Val\`encia}
\affil[ ]{\texttt{\{skondylatos, iprapas, ipapoutsis\}@noa.gr}}
\affil[ ]{\texttt{\{gustau.camps\}@uv.es}}

\begin{document}

\maketitle

\begin{abstract}

We introduce Mesogeos\footnote{Inspired from the Greek word $\mathrm{\textrm{M}}\varepsilon\sigma\acute{o}\gamma\varepsilon\iota\textrm{o}\varsigma$ (Mes\'ogeios), widely adopted to refer to the Mediterranean Sea.}, a large-scale multi-purpose dataset for wildfire modeling in the Mediterranean.
Mesogeos integrates variables representing wildfire drivers (meteorology, vegetation, human activity) and historical records of wildfire ignitions and burned areas for 17 years (2006-2022). 
It is designed as a cloud-friendly spatio-temporal dataset, namely a datacube, harmonizing all variables in a grid of 1km x 1km x 1-day resolution. 
The datacube structure offers opportunities to assess machine learning (ML) usage in various wildfire modeling tasks. 
We extract two ML-ready datasets that establish distinct tracks to demonstrate this potential: (1) short-term wildfire danger forecasting and (2) final burned area estimation given the point of ignition. 
We define appropriate metrics and baselines to evaluate the performance of models in each track. 
By publishing the datacube, along with the code to create the ML datasets and models, we encourage the community to foster the implementation of additional tracks for mitigating the increasing threat of wildfires in the Mediterranean.
% to further contribute to data-driven wildfire modeling.
% to foster the implementation of additional machine learning tracks for mitigating the increasing threat of wildfires in the Mediterranean.

% By publishing the datacube, along with the code for generating machine learning datasets and models, we encourage the scientific community to further contribute to data-based wildfire modeling.

% In the context of climate change that is expected to exarcebate wildfire frequency and intensity in the highly vulnerable Mediterranean region, 
% Additionally, we present the opportunities that Mesogeos offers by suggesting other tasks, like wildfire susceptibility, prediction of extreme wildfires, or burned area size prediction.

% In addition, Mesogeos presents novel prospects for leveraging machine learning in various wildfire modeling tasks, including wildfire susceptibility assessment, extreme wildfire prediction, and burned area size estimation. We release the dataset, accompanied by the corresponding code and machine learning tracks and models, with a strong call to action for the research community to actively contribute to the advancement of our understanding and anticipation of wildfires. We particularly encourage the development of additional tracks and models that leverage the capabilities of the Mesogeos data cube, thereby fostering comprehensive improvements in wildfire analysis.

\end{abstract}

\section{Introduction}

% \footnotetext{Inspired from the Greek word $\mathrm{\text{M}}\varepsilon\sigma\acute{o}\gamma\varepsilon\iota\text{o}\varsigma$ (Mes\'ogeios), widely adopted to refer to the Mediterranean Sea.}

Wildfires play a key role in the ecosystem \cite{pausas_flammability_2017, bond_fire_2005, doi:10.1073/pnas.0604090103, doi:10.1126/science.1163886}, yet they present risks to both humans and the environment \cite{pettinari_fire_2020}.
The threat is inflated by climate change, which aggravates the frequency and extremity of wildfire events \cite{pausas_wildfires_2021}, particularly in Mediterranean-type climate regions \cite{moreira_wildfire_2020, batllori_climate_2013}. 
The changes are expected to be more and more prevalent in the following years \cite{klausmeyer_climate_2009}; thus, there is a pressing need for innovative solutions to enhance wildfire preparedness and management, enabling adaptation to evolving conditions.
The development of such solutions is hampered by the complexity to model wildfires, resulting from the dynamic interactions between several fire drivers such as climate, vegetation, and human activity \cite{archibald_what_2009}, operating across different spatial and temporal scales. 
% These drivers operate across different spatial and temporal scales, making modeling wildfires a very challenging process. 

% Traditionally, the dynamics of fire are modeled using physical models. 
% \cite{van1974structure} is used operationally to evaluate fire danger. 
% Similarly, physics-informed models predict fire spread \cite{Andrews_1986, Finney_1998} as a function of heat flux and fuel availability, ignoring the interdependencies among other influencing factors. 
% While these approaches are widely used worldwide, modeling wildfires can be enhanced by data-driven strategies considering all relevant covariates and their interactions.

% Despite the importance of modeling the interactions between all fire drivers when modeling fires, many applications ignore them
% For example, wildfire danger is most often predicted using meteorologically derived fire weather indices. 
% Particularly, in Europe, the Fire Weather Index (FWI) is used operationally \cite{van1974structure} to evaluate fire danger. 
% Similarly, fire spread is predicted by physics-informed models \cite{Andrews_1986, Finney_1998} as a function of heat flux and fuel availability, that ignore the interdependencies among other influencing factors.

Traditional models \cite{van1974structure, Andrews_1986, Finney_1998} ignore these intricate interactions.
In contrast, Machine Learning (ML) offers the potential to capture them in a data-centric manner.
Nevertheless, the application of ML in the context of wildfires requires careful consideration \cite{prapas_deep_2021}. 
The wildfire occurrence is stochastic, which means that the same environmental conditions may lead or not to a fire ignition. 
Moreover, wildfires are rare events that can lead to imbalanced or sparse datasets. 
Despite the challenges, ML has been employed successfully in several applications \cite{jain_review_2020}.
Particularly, Deep Learning (DL) has been suggested as a method for modeling Earth System problems, including wildfires \cite{reichstein_deep_2019, CampsValls21wiley}. 

Although the potential of DL in wildfire modeling appears promising, its adoption is still not widespread. 
One major obstacle is the limited availability of extensive datasets necessary to support its utilization. 
The vast amount of data required to model wildfires at a larger scale presents difficulties in the collection and curation of the data.
The data sources are often scattered across different platforms and become available in diverse formats and resolutions.
% Additionally, the sheer volume of data analyzed demands preprocessing to create a cohesive dataset. 
Thus, the community lacks a large-scale dataset suitable for various ML tasks in the context of wildfires.

In this work, we introduce \textit{Mesogeos}, an extensive multi-purpose dataset designed to support the development of ML models for various wildfire applications in the Mediterranean.
% To the best of our knowledge, Mesogeos is the first open-access wildfire-related dataset, covering the fire-prone Mediterranean region.
It contains a complete set of variables associated with fire drivers, i.e. meteorological conditions, vegetation characteristics, and anthropogenic factors. 
It also encompasses past burned areas, fire ignition points, and burned area sizes that can serve as predictands for diverse ML tasks.
Mesogeos is harmonized in a standard spatiotemporal grid format, namely a datacube \cite{mahecha_earth_2020}, with a daily temporal resolution and a spatial resolution of $1km\times 1km$, containing data from $2006$ to $2022$.
The datacube structure facilitates the extraction of ML-ready datasets for numerous applications.
To the best of our knowledge, Mesogeos is the largest harmonized, multi-purpose dataset for data-driven wildfire modeling.
% To enhance the reliability of the products related to the predictands, we implement rigorous pre-processing, before appending the data in the datacube.

To demonstrate the datacube's potential applications, we extract from it two ready-to-consume ML datasets: one tailored for the next day's wildfire danger forecasting and one for burned area size prediction, given the ignition.
We employ DL models to establish benchmarks for the two datasets.
Furthermore, we propose several additional directions for utilizing the dataset, suggesting its capabilities for addressing other wildfire-related applications. 
To encourage further research and facilitate the development of similar datasets, we openly publish the Mesogeos datacube, the derived datasets and models, and the code used to generate them \cite{spyros_kondylatos_2023_7741518}. We also provide a github repository: \href{https://github.com/Orion-AI-Lab/mesogeos}{https://github.com/Orion-AI-Lab/mesogeos} and a website for the project: \href{https://orion-ai-lab.github.io/mesogeos/}{https://orion-ai-lab.github.io/mesogeos/} with information on how to use the data and code.
These resources can be a valuable reference for future implementations and extractions of similar datasets.

\section{Related Work}

% The scope of this section is to present i) the recent developments of DL in wildfire applications and ii) the publicly available fire-related datasets along with their limitations. 

DL has demonstrated successful applications in various tasks related to wildfires. 
\citet{huot_deep_2020} have built segmentation models for predicting fire danger with U-Net-type architectures. 
% Their approach utilized the MOD14A1 product \cite{Giglio2021-di} from the Moderate Resolution Imaging Spectroradiometer (MODIS) satellite as the target variable. 
\citet{radke_firecast_2019} developed FireCast, a fire spread model leveraging Convolutional Neural Networks (CNNs) that demonstrated superior performance compared to physics-based models. 
Similarly, \citet{hodges_wildland_2019} and \citet{burge_convolutional_nodate} employed DL techniques to predict fire evolution by training on fire simulations.
Lastly, \citet{ba_smokenet_2019} addressed the fire detection task, by developing SmokeNet, a CNN-based model that was trained to predict hotspots, as provided by the Moderate Resolution Imaging Spectroradiometer (MODIS) Active Fire (AF) data product \cite{giglio_collection_2016}.
Although these studies showcased the potential of DL in various wildfire applications, the datasets used in each work remain unpublished, thus it is impossible for the community to reproduce or improve the results.

% DL has been applied successfully in various applications related to wildfires, where different fire data products have been utilized as target variables.
% \citet{huot_deep_2020} focuses on building models for predicting fire danger, utilizing the MOD14A1 product \cite{Giglio2021-di} of the Moderate Resolution Imaging Spectroradiometer (MODIS) satellite as the target source, which contains a collection of daily fire mask composites.
% \citet{radke_firecast_2019} model fire spread, by predicting fire perimeters, as provided by firefighters in the field.
% Similarly, \citet{hodges_wildland_2019} and \citet{burge_convolutional_nodate} employ DL techniques to predict fire evolution by training on fire simulations, as produced by \cite{Finney_1998}.
% Lastly, \citet{ba_smokenet_2019} addresses the fire detection task, by developing SmokeNet, a model that is trained to predict MODIS active fire data \cite{giglio_collection_2016}.
% These studies demonstrate the successful utilization of DL in different wildfire-related tasks, emphasizing the diverse sources of fire data used as targets for training DL models.

When it comes to modeling wildfires, many studies rely on satellite-derived data, such as AF products that detect thermal anomalies or burned area products that locate rapid reflectance changes. MODIS and VIIRS satellites offer such openly accessible products and are commonly used due to their high temporal resolution, offering daily global coverage.
The MODIS AF product exhibits a spatial resolution of $1km\times 1km$ and has been generating data since $2002$. 
It operates by employing thermal sensors to identify anomalous thermal signatures associated with ongoing fires \cite{giglio_collection_2016}.
The VIIRS satellite, introduced in $2012$, follows a similar AF detection methodology for fires, holding an enhanced spatial resolution of $375m\times 375m$, which leads to a better response to relatively small fires and possesses an improved nighttime performance \cite{schroeder_new_2014, csiszar_active_2014}.
Several studies have been undertaken to assess the quality of these products by comparing their outcomes against human-collected fire databases. 
These analyses have brought to light certain limitations associated with their utilization for the assessment of wildfires.
In the United States and China, MODIS demonstrated a moderate level of concurrence with actual fire data \cite{fusco_detection_2019, fornacca_performance_2017}.
Moreover, both MODIS and VIIRS products exhibited disagreements when evaluated against real fire occurrences in Turkey, a Mediterranean-type region, with more favorable results observed for larger fires \cite{coskuner_assessing_2022}.

% Alternative datasets include MOD14A1 \cite{Giglio2021-di}, an open-source product derived from the MODIS satellite, containing a  collection of daily fire mask composites.
% Despite its good spatial resolution of $1km\times 1km$ and its rapid availability, a study conducted by \citet{masocha_accuracy_2018} demonstrated a poor index of agreement between MOD14A1 outcomes and groundfire data in a  region of Africa.
% \textcolor{red}{Moreover, the MCD64A1 burned area product, maps the spatial extent and approximate date of biomass burning worldwide at a spatial resolution of $500$m \cite{giglio_collection_2018}.
% This product has exhibited inaccuracies in mapping cropland burned areas in Ukraine \cite{hall_validation_2021}, savannas in Brazil \cite{rodrigues_how_2019}, as well as across over six vegetation types in boreal Eurasia \cite{zhu_size-dependent_2017}. 
% Notably, significant underestimation of small-scale fires was observed in these cases.}
Alternative datasets sourced from MODIS include MOD14A1 \cite{Giglio2021-di} and MCD64A1 \cite{giglio_collection_2018}. The former is an open-source product, containing a  collection of daily fire mask composites at a spatial resolution of $1km\times 1km$. The latter, becomes available at a spatial resolution of $500m\times 500m$, mapping the spatial extent and approximate date of biomass burning worldwide. Several validation studies have been undertaken to assess the accuracy of these datasets, revealing instances of disagreement between their outputs and reliable fire records \cite{masocha_accuracy_2018, hall_validation_2021, rodrigues_how_2019, zhu_size-dependent_2017}.
Apart from MODIS products, there are several other publicly available fire datasets derived from Earth Observation satellites.
These include global datasets such as FRY \cite{laurent_fry_2018} and Fire Atlas \cite{andela_global_2019} which are designated to deliver the total burned area of fires and GlobFire \cite{artes_global_2019} which provides daily fire perimeters. 
In Europe, the European Forest Fire Information System (EFFIS) \cite{san2003european} provides accurate burned area estimates following a semi-supervised approach that uses different satellite sensors, estimating about $95\%$ of the total area that burns in Europe every year \cite{effis_website}. The EFFIS burned area product is used in this work because of its improved accuracy in the Mediterranean region. 

While these datasets provide only fire data, a comprehensive wildfire analysis and modeling needs to incorporate variables related to fire drivers, such as vegetation, weather, drought, and topography. 
In this direction, several studies have published datasets that integrate fire targets with variables related to fire ignition and spread.
These datasets often have limitations, such as focusing on specific small-scale regions or exhibiting coarse spatial and temporal resolutions. 
Furthermore, they are typically designed for specific tasks tailored to a single ML objective. 
\citet{kondylatos_wildfire_2022} have published a dataset covering Greece, which is specifically designed for forecasting the next day's wildfire danger. 
Though they also publish a datacube, its applications are limited by its small size, only covering a part of the eastern Mediterranean.
\citet{huot_next_2022, singla_samriddhi_wildfiredb_2021, diao_uncertainty_nodate} have introduced datasets designated for wildfire spread prediction in the continental US.
The former relies on the MODIS AF product as the target variable, while the others utilize the VIIRS AF product.
Moreover, \citet{sayad_predictive_2019} have shared a dataset tailored to wildfire modeling in a small region of Canada, recording a limited number of fire events.
Finally, \citet{prapas_deep_2022} presented a global dataset for seasonal fire danger forecasting, but in a coarse spatial and temporal resolution. 
% In contrast to the existing datasets, Mesogeos is a large-scale dataset, that offers a good spatial and temporal resolution and is designed to cater to multiple ML objectives, thereby paving the way for significant advancements in data-driven wildfire modeling.
\paragraph{Comparison with Existing Datasets and Contributions.} Mesogeos is a large-scale, versatile, multi-purpose dataset designed to cater to a multitude of ML tasks related to wildfire modeling.
It sets itself apart from datasets \cite{kondylatos_wildfire_2022, huot_next_2022, singla_samriddhi_wildfiredb_2021, diao_uncertainty_nodate, sayad_predictive_2019}, which focus solely on single ML tasks.
It offers a broad scope, by encompassing a wide range of daily inputs covering many relevant fire drivers across the entire area of interest. 
It is provided in a cloud-optimized datacube structure, offering spatio-temporal metadata, that uniquely associates data points with specific date, longitude, and latitude values.
This structure empowers researchers to select subsets in any dimension, retrieve variables, calculate new ones, or even augment the datacube with other data.
Such inherent flexibility simplifies data access, facilitates the extraction of diverse ML datasets, and enables the expansion of the dataset.
Thus, it allows its adaptation to a wide range of ML applications, based on individual research needs. 
Moreover, by leveraging the EFFIS burned area product while refining the provided ignition dates of fires through cross-comparison with the MODIS AF product (as illustrated in Section \ref{sec:3}), we achieve a more reliable representation of burned areas and fire ignitions compared to existing datasets \cite{giglio_collection_2016, schroeder_new_2014, Giglio2021-di, giglio_collection_2018, laurent_fry_2018, andela_global_2019, artes_global_2019, toulouse_computer_2017}.
Finally, it is worth noting that Mesogeos is the first dataset of this resolution and scale tailored for wildfire modeling in the Mediterranean region, a fire-prone area that has lacked dedicated datasets of this nature.

\section{Mesogeos Datacube}\label{sec:3}

\paragraph{General information.} The Mesogeos dataset is structured as a spatio-temporal datacube with three dimensions: longitude, latitude, and time. 
The datacube encompasses $27$ variables related to meteorology, vegetation, land cover, and human activity.
All these data are well-known fire drivers and can be used as predictors in wildfire-related applications.
Mesogeos also includes historical burned areas, ignitions, and burned area sizes as separate variables.
It has $1km\times1km\times daily$ resolution and contains the values of the variables covering the period from $2006$ to $2022$.
It incorporates data from the wide Mediterranean area and spans a total area of $4714km\times 1753km$ and $6026$ days. 
% Depending on the application, these variables can be used as either predictands or predictors.
% The source, original spatial and temporal resolution, and the units of each variable are presented in Table \ref{tab:variables}.

\paragraph{Data Sources.} 
\citet{hantson_status_2016} study the complex interactions between the variables that control fire.
They divide fire controllers into three main categories: human, weather, and vegetation.
Weather conditions and vegetation play a crucial role in determining the rates of fuel drying and therefore affecting fire occurrence and spread. 
Topography also influences fire behavior as fire fronts travel faster uphill because of the upward convection of heat.
While natural factors such as weather, vegetation, and fuel load impact fire occurrence, human activity-related outcomes such as intentional or accidental fire ignition, land conversion, and population density also significantly shape fire regimes.
In this study, in an attempt to cover all the factors influencing fire occurrence, we collect data sources containing information about all the aforementioned fire drivers.

The meteorological data (temperature, wind speed, wind direction, dewpoint temperature, surface pressure, relative humidity, total precipitation, surface solar radiation downwards) are collected from the ERA5-Land database \cite{munoz2021era5}, which contains historical hourly land weather measurements from $1950$ to today. 
We use day's and night's land surface temperature \cite{wan_mod11a1_2018}, Normalized Difference Vegetation Index (NDVI) \cite{didan_mod13a2_2015}, and Leaf Area Index (LAI) \cite{myneni_mod15a2_2002} from MODIS and soil moisture index from the European Drought Observatory (EDO) \cite{cammalleri_comparing_2017}. 
These data are used as proxies for the vegetation status and drought.
Distance from roads and population are downloaded from Worldpop \cite{tatem2017worldpop} and are used as indicators of human activity. At the same time, topography data, i.e. elevation, slope, aspect, and curvature are gathered from the Copernicus DEM - Global Digital Elevation Model (COP-DEM) \cite{European_Space_Agency2022-nc}.
The land cover classes are collected from the Copernicus Climate Change Service \cite{Copernicus_Climate_Change_Service2019-lw}. 
The burned areas come from EFFIS.
Finally, MODIS AF product \cite{giglio_collection_2016} is used to estimate ignition cells and the ignition date. 

\begin{figure}
    \centering
    \includegraphics[width=\linewidth]{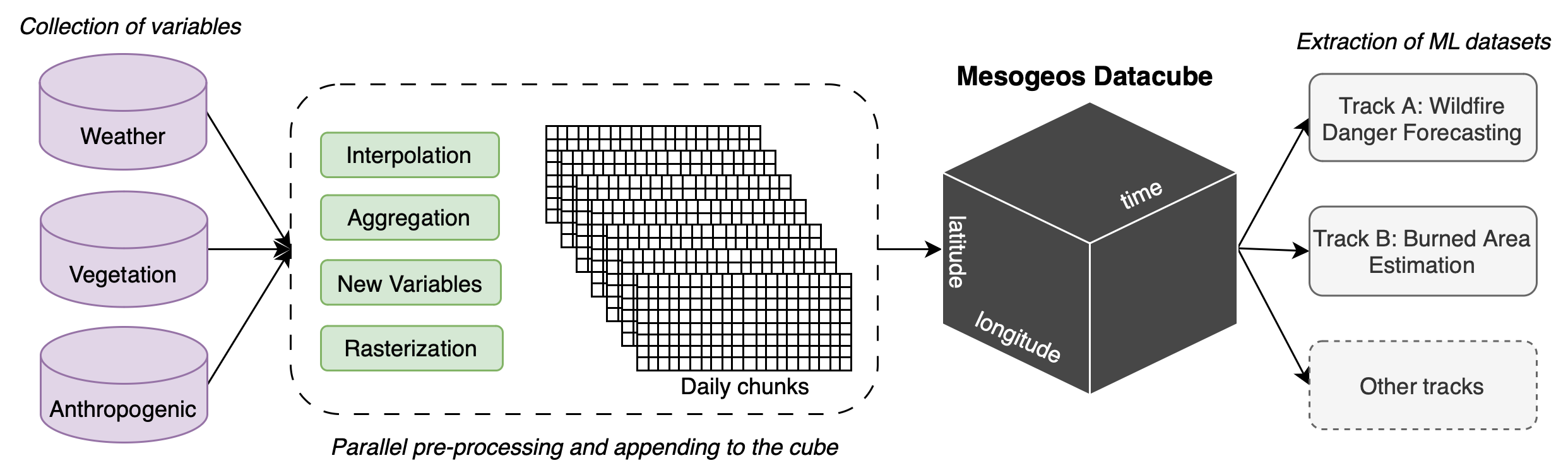}
    \caption{Pipeline of the datacube construction. The data are collected from various sources. The data inputs are pre-processed using interpolation, aggregation, calculation of new variables, and rasterization and the final daily chunks are appended in the datacube on the corresponding date. This process runs in parallel for multiple days to enhance the efficiency of the process. After the creation of the datacube, the ML datasets are extracted from it.}
    \label{fig:pipeline}
\end{figure}

\paragraph{Datacube Creation.} Creating a unified dataset that stores all wildfire-related information in a standard format will later permit an easy extraction of ML-ready datasets for different tasks.
Consequently, we have opted to gather and harmonize all the data into a spatio-temporal datacube format to leverage this structure's various capabilities regarding easy access, manipulation, and extraction of data.
Nevertheless, the creation of such a datacube poses significant challenges.
The substantial volume of data (TBs of unprocessed data) necessitates significant downloading and storage capabilities. 
Additionally, the data are sourced from different providers, each with their own access points and formats (e.g. vector or raster), making data acquisition a challenging task. 
Furthermore, the original resolution of each variable varies, requiring harmonization to match the expected resolution of the datacube. 
Consequently, the construction of the datacube demands careful and efficient development for minimizing time and resource requirements.

To address the challenges above, we create the pipeline illustrated in Figure \ref{fig:pipeline} for the creation of the datacube. We follow these steps:
Firstly, we collect and store all relevant variables from the various input sources.
Subsequently, we construct the datacube's structure by generating a grid of dimensions $1km \times 1km \times 1-day$ and defining daily chunks. 
The daily chunking applies independently to each variable, which means that values are stored in different files for each day.
Finally, we append each input source to the datacube on a day-to-day basis to prevent memory errors.
For each day, we perform all the necessary pre-processing steps, such as converting data into raster format, conducting temporal or spatial interpolation/aggregation, changing coordinate systems, and doing variable calculations.
Then, we store the values in the chunk that refers to the corresponding date. 
As chunks are stored independently, we make this process totally in parallel. 
We use the xarray \cite{hoyer2017xarray} python package for development. 
Using the default Zarr \cite{Miles2020-ad} compression, the datacube occupies a storage space of $648$ GB, while the memory needed to load the datacube is much larger, at around $3.2$ TB, assuming $32$-bit floats for the dynamic variables. 
The code for creating this datacube is made available and can be consulted to further enhance the existing dataset.
Notably, this pipeline can be adapted with minor adjustments for generating similar datacubes applicable to various Earth science domains.
For a more comprehensive understanding of the pre-processing procedures undertaken for each variable, please refer to the Supplementary Material.

% \begin{enumerate}
%     \item We collect and store all the variables in an internal machine.
%     \item We create the structure of the datacube, by constructing a $1km\times1km$ grid cell for all the days of interest with daily zero values for all the variables.
%     \item We append each input source serially in the datacube on a day-to-day to avoid memory leaks.
%     Specifically, for each day we make all the needed to pre-process (.e.g. turn into raster format, interpolation in temporal or spatial dimensions, change of coordinate system, calculation of variables ) and we store the daily values of the variables in the empty datacube for the correct date.
%     This process runs for many days in parallel for better efficiency.
% \end{enumerate}

\paragraph{Burned Areas Dataset.} The burned areas dataset provided by EFFIS is an improvement over the MODIS products, offering a more reliable and credible resource. 
EFFIS enhances the burned area data obtained from MODIS by employing a semi-supervised processing of imagery from various satellites, i.e. Sentinel-2, and VIIRS. 
This process involves semi-automatic procedures aimed at enhancing the quality of fire maps \cite{effis_website}.
% It is important to acknowledge that the datacube, which contains burned areas from $2006$ to $2022$, has certain limitations concerning the availability of burned areas data.
Despite its improved quality, it is important to note that the dataset does not offer the same coverage of burned areas across all Mediterranean countries throughout the specified timeframe, resulting in the absence of data for certain countries in specific years.
This is further analyzed in the Supplementary Material.
% Therefore, it is crucial to recognize that the availability of burned area information is not uniform across all countries over the entire study period.
% Nevertheless, t
% Initially, fires are identified through an unsupervised procedure that incorporates band thresholds, ancillary data, active fire detection, and fire news applications. Subsequently, the detected fires undergo visual verification and correction through visual interpretation.
% Thus, while acknowledging the limitations associated with the burned area data across countries and years, it is important to recognize the improved quality and credibility of this dataset.
% However, it is important to say that this is the best available dataset in the Mediterranean. 
% If better data are found they can be easily integrated in the datacube

 \paragraph{Ignition Date Calculation.} The start dates of the fires provided by EFFIS may not always correspond to the date of ignition of the fire \cite{effis_website}.
However, the accurate calculation of the first detection of a wildfire is extremely important in order to avoid data leakage and enhance the transparency and precision of the training process.
% Thus, further pre-processing is necessary to determine the precise ignition date of each wildfire.
For this, we implement a method that involves intersecting burned areas from EFFIS with AF obtained from MODIS.
We use each burned area identified by EFFIS as a representative instance of a distinct fire event.
We then select hotspots from a $1km$ spatial buffer zone surrounding the burned area and a temporal buffer of $7$ days around the date of the ignition as provided by EFFIS.
From this selection, we identify the hotspot with the oldest date within the buffer as the ignition point of the fire and its date as the ignition date of the fire.
We discard the specific wildfire from the dataset if no hotspots are detected within the designated buffer zone. 
Notably, MODIS AF are used solely for ignition date refinement and ignition point identification and are not employed as primary anchors for fire events.
Despite the improvements achieved through our approach, it is essential to recognize that some misalignments in the ignition dates may persist.

\section{Machine Learning tracks}\label{sec:4}

% We devise two tracks for the use of ML algorithms with Mesogeos. 
% Next, we discuss further applications and possibilities that Mesogeos provides.

\subsection{Track A: Wildfire danger forecasting}\label{sec:4a}

\paragraph{Task formulation.} 
% Following \cite{kondylatos_wildfire_2022}, we define fire danger for a given cell and a given day $t$, as the probability of a fire occurring on the day $t$ and becoming large, given the values of the different fire drivers $x_t$ in the preceding days. 
% We assume wildfires exceeding $30$ hectares indicate high wildfire danger, while low wildfire danger is associated with the absence of any wildfire within a specified buffer zone surrounding a given pixel.
% However, in the current work, we introduce a slight modification to the assessment of fire danger compared to \cite{kondylatos_wildfire_2022}.
% Other than including all the pixels of the burned areas as indicators of high danger, we treat the ignition point as a reference point of the event and the final burned area size as a measurement of the corresponding fire event's danger level.
% Thus, we interpret a more significant fire expansion as indicating greater danger.
% This alteration in the fire danger assessment is motivated by two reasons: i) the vast amount of data hampers the sampling of all burned area pixels, and ii) there is significant auto-correlation among pixels within the same fire event, imitating the sampling of multiple instances of the same pixel.
% The task is formulated as a time-series binary classification problem.
For a given cell and a given day $t$, we define fire danger as the probability of a fire occurring on the day $t$ and becoming large, given the values of the different fire drivers $x_t$ in the preceding days. 
We assume that a wildfire exceeding $30$ hectares indicates high wildfire danger.
To measure this danger, we treat the ignition point of the fire as a representative point of the event and the final burned area size resulting from it as an indicator of the corresponding danger level.
Conversely, low wildfire danger is associated with the absence of any wildfire within a specified buffer zone surrounding a given pixel. 
In alignment with prior research in the field of data-driven wildfire danger forecasting \cite{huot_deep_2020, kondylatos_wildfire_2022, zhang_forest_2019}, the task is formulated as a binary ML classification problem.
One class signifies increased danger, while the other represents low-danger instances.
However, in this work, we slightly modify the standard classification loss, involving a weighting scheme that considers the burned area size of each distinct event.
This modification aims to interpret a more significant fire expansion as an indication of higher danger.
The resulting softmax probabilities of the trained classification model serve as indicators of the level of fire danger.

\paragraph{Dataset extraction.} 
We extract a time-series dataset, consisting of days $t-1, t-2,..., t-30$ of the dynamic input observations and the static features repeated in time.
Positive class examples consist of the ignition points of the fires that started on the day $t$.
% As wildfire occurrence is stochastic the lack of a fire event does not correspond necessarily to a lack of fire danger.
% Thus, the sampling of negative cells is a challenging task as there is the danger of sampling negatives that actually imply high fire danger.
% To mitigate this danger, f
For the negative class examples, we select cells outside a buffer of $62km$ from any fire that started on this day to mitigate the risk of choosing cells that imply great danger but did not burn.
Moreover, we follow the sampling strategy as in \cite{kondylatos_wildfire_2022} and sample i) two times more negatives than positives, ii) the negatives following the land cover distribution of the positives.

\paragraph{Experimental Setup.}
For the experiments, we use a Long Short-Term Memory (LSTM) architecture \cite{hochreiter1997long} and the encoder of a Transformer model \cite{vaswani_attention_2017} as standard models for time-series data.
Moreover, we employ a Gated Transformer Network (GTN) \cite{liu_gated_2021} that uses the attention mechanism both in time and in variables, which could aid in modeling the complex interactions of the variables in the current task. 
The models are optimized using the cross-entropy (CE) loss.
To let the ML models learn to assign greater danger to larger fires, we weigh the loss based on the size of the wildfire's burned area. 
For this, we multiply the standard CE loss value of a given sample by the corresponding burned area size resulting from it. 
In practice, in order to prevent the larger fires from totally dominating the learning process, we apply a logarithmic transformation to the burned area size multiplier, in an attempt to narrow the penalization gap between small and large fires.
Negative samples are assigned weights equal to the minimum burned area size among the positive samples. 
This ensures that the negatives, representing low-danger instances, receive adequate attention during the training process, but not more than any high-danger instance.
The hyperparameters for each model are tuned separately using the validation set.
The years $2006-2019$ are used for the training set, $2020$ is used as the validation set, and the years $2021-2022$ are used as a test set. 
The final dataset consists of $25722$ samples ($8574$ positives and $17148$ negatives), from which $19353$ ($6451$ positives and $12902$ negatives) are in the training set, $2262$ ($754$ positives and $1508$ negatives) are in the validation set and $4107$ ($1369$ positives and $2738$ negatives) are in the test set.
All the available input variables from the datacube are used in the experiments.
They are all normalized before passing into the model.
Precision, Recall, and Area Under Precision-Recall Curve (AUPRC) are used as metrics for the evaluation of the performance of the models.
The details about the architectures of the models and the hyperparameters are provided in the Supplementary Material.

\begin{table}[htbp]
    \caption{Results of the fire danger forecasting track}
    \centering
    \begin{tabular}{ccccc}
        \toprule
        Model & Precision & Recall & $F1$ & AUPRC \\
        \midrule
        LSTM \cite{hochreiter1997long} & 0,763 & \textbf{0,812} & \textbf{0,786} & 0,853 \\
        Transformer \cite{vaswani_attention_2017} & \textbf{0,802} & 0,759 & 0,780 & 0,856 \\
        GTN \cite{liu_gated_2021} & 0,781 & 0,790 & \textbf{0,786} & \textbf{0,858} \\
        % TCN & 0,757 & 0,803 & 0,779 & 0,847 \\
        % InceptionTime \\
        % gMLP & 0,772 & 0,801 & 0,786 & 0,848 \\
        % XceptionTime \\
        % ResCNN & 0,789 & 0,788 & 0,788 & 0,860 \\
        % FCN & 0,788 & 0,777 & 0,782 & 0,863
        \bottomrule
    \end{tabular}
    \label{tab:results_danger}
\end{table}

\begin{figure}[htbp]
    \centering
    \includegraphics[width=\linewidth]{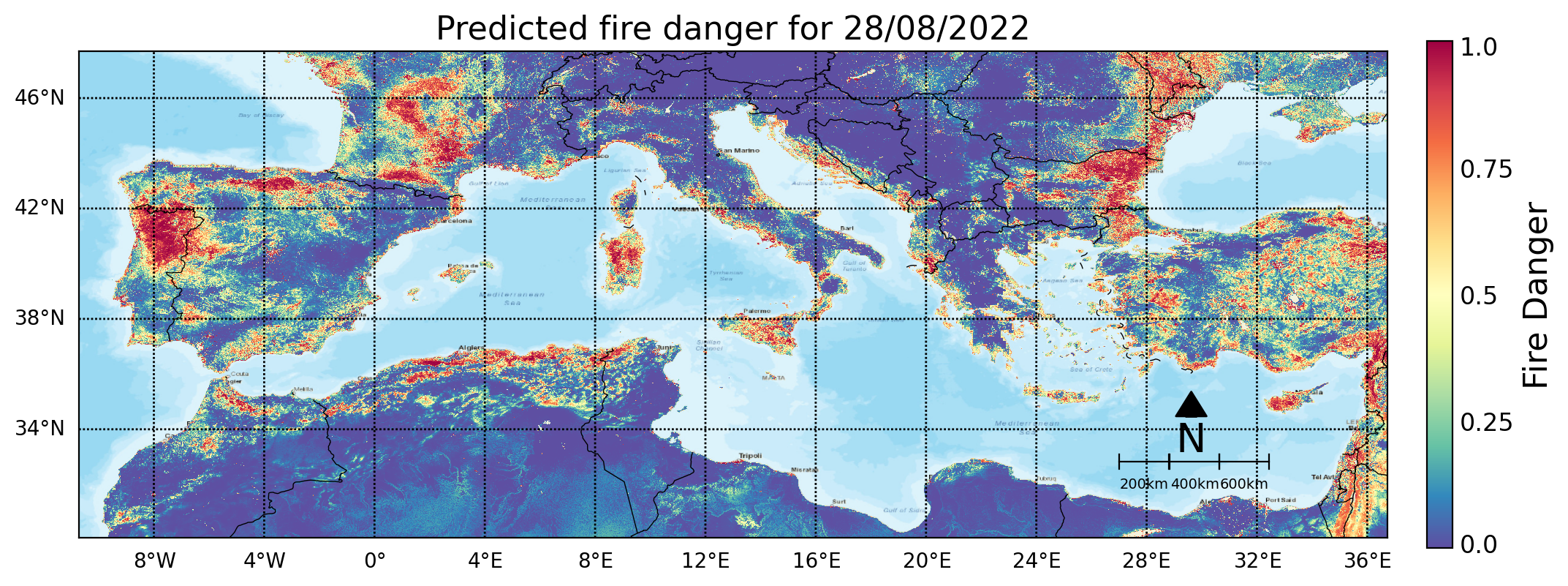}
    \caption{A wildfire danger map of the Mediterranean, produced by the Transformer model. The fire danger is indicated by the softmax probabilities of the trained model.}
    \label{fig:map}
\end{figure}

\paragraph{Results.} The results of the models are shown in Table \ref{tab:results_danger}. 
The promising results from all the models show that they can distinguish between high and low fire danger instances.
It should be noted that the optimal performance model varies for each metric, thus making it difficult to define the overall best-performing model.
In addition to the quantitative metrics, we also provide a visualization map generated by the Transformer model, presented in Figure \ref{fig:map}. 
The map displays the model's softmax probabilities, indicating fire danger levels. 
This provides an example of how a daily fire danger map can be generated for operational scenarios in the Mediterranean region. 
The high spatial resolution of the dataset employed in this study could enable improved fire management strategies and demonstrate potential operational benefits.
In this example, concerning fire danger for a day in the summer of $2022$ notable variations in fire danger are observed across different areas of the Mediterranean, as well as within individual regions of each country. 
% More fire danger maps are provided on the website.

\subsection{Track B: Final Burned Area Prediction}\label{sec:4b}

\paragraph{Task formulation.} This track focuses on predicting the likely extent of a wildfire's final burned area, given the ignition point and a set of variables available at the time of ignition, representing the fire drivers inside a neighborhood around the ignition point.
These fire drivers encompass factors that influence fire behavior and spread.
Thus, the objective of the ML task is to estimate the likelihood of the neighboring pixels surrounding the ignition point, to be eventually contained within the final burned area of the wildfire.
The final burned area prediction is treated as a segmentation task, with two classes, indicating whether a pixel will experience burning or remain unaffected by the ignited fire.
The resulting softmax probabilities of the trained model serve as indicators of the likelihood of a pixel being burnt.

\paragraph{Dataset extraction.}
For every fire event, we extract $64km\times64km$ patches, that are centered around the fire's ignition point, usually containing the whole burned area of a given fire event. The extracted samples include all the values of the variables of the datacube for the date of the fire's occurrence.
% The selection of a spatial buffer of $64\times64$ is selected in an attempt to capture the full extent of the majority of fires of the datacube. 
% We decide not to sample an even bigger spatial buffer, to avoid an inflation of the dataset size and the inclusion of numerous negative instances, which would lead to an even more imbalanced dataset and hamper the training process.

\paragraph{Experimental Setup.} 
The $64\times64$ patches are randomly cropped to $32\times32$ during the training process.
This approach ensures that the ignition point remains within the patch while preventing the model from generating a bias towards fire expansion solely from the central cell. As this is a segmentation task, we use the U-Net architecture \cite{ronneberger_u-net_2015} with an EfficientNet-B1 \cite{tan2019efficientnet} encoder. Different input variables are stacked as separate channels. The cross-entropy loss is used to train the models' parameters.
To define a baseline for the task, we train an additional model that uses as input only the ignition points. 
We do a temporal split to avoid leaking data from fire events happening close in time, using $2006-2019$ for training ($12550$ samples), $2020$ for validation ($1781$ samples), and $2021-2022$ for testing ($3527$ samples). The loss in the validation is used for early stopping. Input variables are scaled with the minimum, and maximum values in the range $[0, 1]$ before being served as inputs to the model.
As evaluation metrics, we report the CE loss and the AUPRC.
The exact architecture and the values of the hyperparameters are provided in the Supplementary Material.

\begin{table}[htbp]
    \caption{Results of the final burned area prediction track}
    \label{tab:results_spread}
    \centering
    \begin{tabular}{ccccc}
        \toprule
        Model & CE Loss & AUPRC \\
        \midrule
        U-Net (only ignitions) & 0.0177 & 0.394 \\
        U-Net (all variables) & 0.0166 & 0.418 \\
        \bottomrule
    \end{tabular}
\end{table}

\begin{figure}[htbp]
    \centering
    \includegraphics[width=\linewidth]{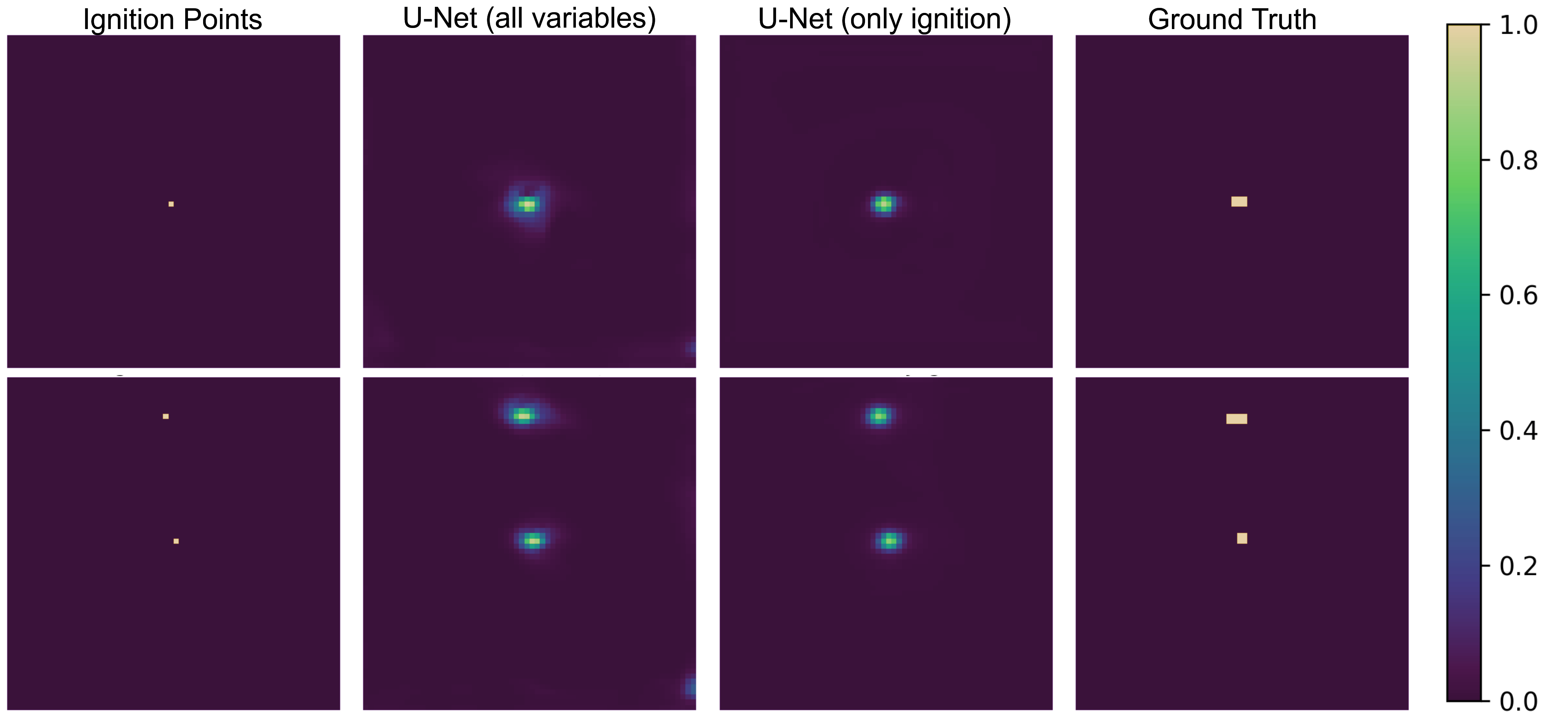}
    \caption{Two examples of predictions (softmax values for the positive class) from the U-Net using \textit{all variables}, and the U-Net using \textit{only ignition} points. The predictions are presented together with binary maps representing the initial \textit{ignition points} and the \textit{ground truth} burned areas.}
    \label{fig:track_b_figure}
\end{figure}

\paragraph{Results.} The experimental results are presented in Table \ref{tab:results_spread}, which shows that the model incorporating all variables slightly outperforms the baseline model relying only on the ignition as input. 
% This shows that the task is quite challenging and probably needs additional data helpful for the evolution of the fire, like the time series of the variables in the days preceding and following the fire, as in \cite{coffield_machine_2019}. 
% % The track is accompanied by code to extract such datasets. 
% In addition to the quantitative metrics, we showcase some examples of predictions of the model in Fig. 
In Figure \ref{fig:track_b_figure} we see two examples from the models' predictions in the test set.
It is notable that models predict higher spread danger around the ignition point, with the model that uses input from all  variables showing some enhanced skill compared to the no-skill baseline.
% These findings emphasize the importance of further exploring and developing models to understand better the interactions between fire drivers that lead to the resulting burned areas. 
To improve on the task, it might be important to include variables before and after the fire starts and not only from the date of ignition. 
Notably, \citet{coffield_machine_2019} find that the performance can be improved using weather data forecasts for $1-5$ days after the start of a fire. 
The dataset extraction script that accompanies the track includes the capabilities to extract datasets that include time series of arbitrary length before and after the start of each fire.

\section{Potential other tracks to explore}\label{sec:5}

% This section presents a range of potential other tasks that can be investigated using Mesogeos.
Mesogeos incorporates burned areas and their sizes, as well as ignition points as predictands, and thus can be used for various ML applications. 
Thus, users can address additional tracks beyond those presented in the current study. 
The following ideas serve as suggestions for potential paths of research.

\paragraph{Fire size prediction.} This application involves predicting the final size of fires rather than the presence of burned areas, which was addressed in this work. 
This can be framed as either a regression or a multi-class classification task.
As a regression task, the target variable would be the burned area in hectares and therefore predictive models could be developed to estimate the final size of fires.
As a classification task, a model could be employed to categorize fires into small, medium, or large. 
Notably, the dataset initially created for the wildfire danger forecasting track can be used as is for these two tasks.

% \textbf{Fire danger forecasting.} In this work, we constructed the training dataset by sampling ignition points from each fire event along with non-fire pixels for the fire danger forecasting task. 
% To effectively train our machine learning model, we employed a weighting scheme that accounted for the size of each fire, ensuring that larger fires contributed more significantly to the loss function.
% Alternatively, we propose an approach for fire danger prediction, focusing on creating a model that can predict whether a pixel belongs to a burned area. 
% This can be formulated as a segmentation task with two distinct classes: burned and non-burned. 
% Under this approach, pixels within a burned area would be classified as belonging to the burned class, while all other pixels would be assigned to the non-burned class.
% To facilitate adopting this alternative approach, we have provided the necessary code to generate such datasets as part of the codebase for Track A.

% \textbf{Fire ignition prediction.} Fire ignition prediction refers to the task of predicting the location where a fire is likely to initiate. 
% It is crucial to acknowledge that fire ignitions are stochastic events primarily caused by human-related factors, making them very challenging to model accurately. 
% Nonetheless, Mesogeos allows the community to extract datasets associated with fire ignitions and treat them with ML. 
% The dataset generated for Track A can also be utilized without modifications for this particular track.

\paragraph{Extreme events forecasting.} Forecasting extreme events holds significant importance for effective fire management strategies. 
Concerning the Mediterranean, most of the damage caused by wildfires is the result of only a few large fires \cite{san-miguel-ayanz_analysis_2013}.
Therefore, anticipating these wildfires is crucial for ecosystem preservation and optimization of fire management.
Mesogeos includes extreme events, such as the $5$ massive fires in Greece in the summer of $2021$, that burned nearly $94,000$ hectares \cite{giannaros_meteorological_2022}, the extreme events in Portugal in October $2017$, with $5$ events burning more than $18,000$ hectares each \cite{castellnou_fire_2018} and the fire in Valencia, Spain in $2012$ that burned over $50,000$ hectares \cite{vicedo-cabrera_health_2016}.
Extracting a dedicated dataset for this track offers an opportunity to develop models explicitly predicting these extreme events.
As these extreme events are rare, one potential approach to tackle this task would be treating it as an outlier detection problem. 

\paragraph{Wildfire susceptibility mapping.} Wildfire susceptibility is defined as the static probability of wildfires in a certain area, depending on the characteristics of the terrain and prevailing meteorological conditions \cite{trucchia_wildfire_2023}. 
This task can be framed as a binary classification task.
In this context, the dataset extraction process would involve identifying positive samples as the number of all pixels affected by fire over multiple years. 
Negative samples would be obtained from pixels that have never experienced burning, i.e. not belonging to any burned area of the dataset. 
An ML model could then be employed to distinguish between the fire-susceptible and non-susceptible samples.

\paragraph{Self-supervised learning.} The vast amount of data in the datacube remain untapped when extracting task-specific datasets. 
Self-supervised learning (SSL) \cite{balestriero_cookbook_2023} offers a promising approach to take advantage of the full capabilities of these data. 
SSL allows for acquiring a representation that can be utilized across various downstream tasks, including those mentioned earlier.
% However, applying SSL to Earth System problems presents challenges. 
% Other than learning image representations, the model has to learn physical interpretations and relationships between multiple physical variables.
% Consequently, standard SSL techniques may prove insufficient, necessitating the development of alternative methods, specific to this problem.
Concerning the SSL track, extracting specific datasets is unnecessary, as the training samples can be directly extracted from the datacube during the data loading process. 
Careful engineering is essential when selecting and extracting samples to minimize the time required for the model to be trained.
% By employing SSL techniques, researchers can leverage the extensive data in the datacube to improve the performance of various fire-related tasks.

\paragraph{Modeling at different spatio-temporal scales.} Mesogeos has a resolution of $1km\times1km\times daily$, enabling the examination of problems at that specific temporal and spatial scale. However, the flexibility of the datacube format allows for resampling in various temporal and spatial dimensions through appropriate aggregations. This would enable the treatment of other tasks in coarser temporal or spatial scales, like seasonal or sub-seasonal fire modeling.

\paragraph{Beyond traditional ML.} When doing ML-based wildfire modeling for decision support it is many times important to dive deeper into understanding the underlying processes that drive the models' predictions. In that respect, Mesogeos can foster the development of explainable AI techniques \cite{vilone_explainable_2020} toward a better understanding of models and subsequently the interactions of the fire drivers that result in wildfires. Additionally, causal inference methods \cite{scholkopf_towards_2021} could be used to assess the effects of human controls, such as agricultural practices or land use on wildfire regimes. Moreover, considering the stochastic nature of fire processes, noise commonly appears on the labels. Methods that take into account the noisy labels \cite{cordeiro_survey_2020} and especially those estimating the inherent aleatoric uncertainty \cite{collier_simple_2020}, could enhance the reliability of the models and support the decision-making. When existing layers in the datacube are not enough, the datacube can be easily enhanced with extra information such as socio-economic factors, settlement, and infrastructure. 

% One could use the presented dataset to expand with explainable AI and causal inference. Both will require tight collaboration with domain experts.

\section{Limitations}\label{sec:6}

Despite the advantages of Mesogeos, we would like to acknowledge certain limitations. 
Firstly, the datasets inherit inaccuracies of the original data sources. 
Factors such as the satellites' spatial resolution and missing data resulting from cloud cover can influence the precise determination of the fires' location and size. 
Additionally, the acquisition of the fire ignition date is challenging and prone to deviations, as discussed in Section~\ref{sec:3}. 
Another limitation arises from the types of fires included in the burned areas' products.
It is possible that there are fires resulting from prescribed burning or agricultural burning~\cite{effis_website}, which cannot be modeled using just the variables within Mesogeos.
Moreover, while the daily temporal and $1km\times1km$ spatial resolution of the datacube is appropriate for the applications suggested in this research, it cannot be used to address other, nowcasting-type problems related to other cycles of fire management such as fire spread, fire detection, or fire recovery related applications. 
For the target variable, Mesogeos considers the highly reliable final burned areas from EFFIS, but ignores intermediate temporal information of the wildfire evolution, compared to work explicitly targeting wildfire spread \cite{huot_deep_2020, diao_uncertainty_nodate, singla_samriddhi_wildfiredb_2021}.
Furthermore, the dataset lacks information regarding fire suppression efforts. 
The interventions of firefighters and responders have the potential to influence fire dynamics both at the time of ignition, achieved through water application to weaken fire spread, as well as during winter months by means of fuel cleaning or controlled burning.
It should be acknowledged that the absence of such data could influence the modeling of some of the tracks.
For example, when training the model to predict fire danger, it remains unknown whether the fire would have grown larger (indicating higher danger) in the absence of any wildfire suppression measures, or the opposite.
Finally, it is important to highlight that while ML can assist in wildfire modeling, an operational application in wildfire management necessitates a thorough evaluation across fire seasons and against operational baselines, including domain experts and wildfire responders in the process.

% Limitations of ML based wildfires forecasting/modeling. No fine grained info about spread of fire, firefighting etc. Limitations of the burned area dataset (problem with satellite spatial resolution, missing data)

\section{Availability and Maintenance}\label{sec:7}

The Mesogeos datacube and the datasets utilized in this study are made publicly available. The project's website \href{https://orion-ai-lab.github.io/mesogeos/}{https://orion-ai-lab.github.io/mesogeos/} will hold updated links to the data and code repository, as well as a leaderboard for the ML tracks. The repository contains code for generating Mesogeos, extracting datasets for the tracks, and running the models, enabling the reproduction of the results presented in this work. 
We encourage the community to further contribute with more ML tracks and models and advance data-driven wildfire modeling using the Mesogeos datacube.
% TODO talk about code to generate dataset, tracks and benchmarks.

\section{Conclusion}
In conclusion, this work introduces Mesogeos, a valuable resource for data-driven wildfire modeling.
By leveraging the structure of a datacube and incorporating variables that represent various fire drivers and historical wildfires, Mesogeos facilitates the extraction of diverse datasets, empowering researchers to model various fire-related tasks.
In this work, we demonstrate two tracks focusing on fire danger forecasting and burned area prediction to showcase the effectiveness and potential of the dataset.
Lastly, we present several alternative tracks that address a wide range of challenges and tasks associated with anticipating and understanding wildfires, thereby paving the way for new avenues of research and advancement in wildfire modeling.
%Moreover, we present several alternative tracks that Mesogeos offers for addressing a multitude of challenges and tasks associated with wildfires, opening up new avenues for advancing research in wildfire modeling. 

\ack{This work has received funding from the European Union's Horizon 2020 Research and Innovation Projects DeepCube and TREEADS, under Grant Agreement Numbers 101004188 and 101036926 respectively.}

\bibliography{mesogeos_arxiv}
\newpage
\appendix

\section{Supplementary Material}
In the Supplementary Material we:
\begin{itemize}
    \item Provide a detailed description of the dataset.
    \item Provide the details for the deep learning models that were used in the paper.
\end{itemize}

\subsection{Dataset Description}
This subsection provides a detailed description of the dataset.

\subsubsection{Motivation}
Mesogeos is created to enable the development of data-driven wildfire modeling in the Mediterranean. 
Wildfires are one of the most threatening natural phenomena and their impact is expected to aggravate even more due to climate change.
While several physical models are used in various fire-related applications, they do not have the capabilities to model the complex interactions between the different fire drivers, like the data-driven models.
However, the lack of comprehensive datasets hinders the widespread adoption of data-driven modeling in wildfire-related applications.
Besides, even the existing datasets in the domain are restricted to a specific ML application.
Thus, Mesogeos is introduced as a multi-purpose dataset that can be used for the development of several applications.
Moreover, it covers the Mediterranean, one of the most fire-prone regions on Earth, where no other wildfire-related datasets are available.

\begin{figure}[htbp]
    \centering
    \includegraphics[width=\linewidth]{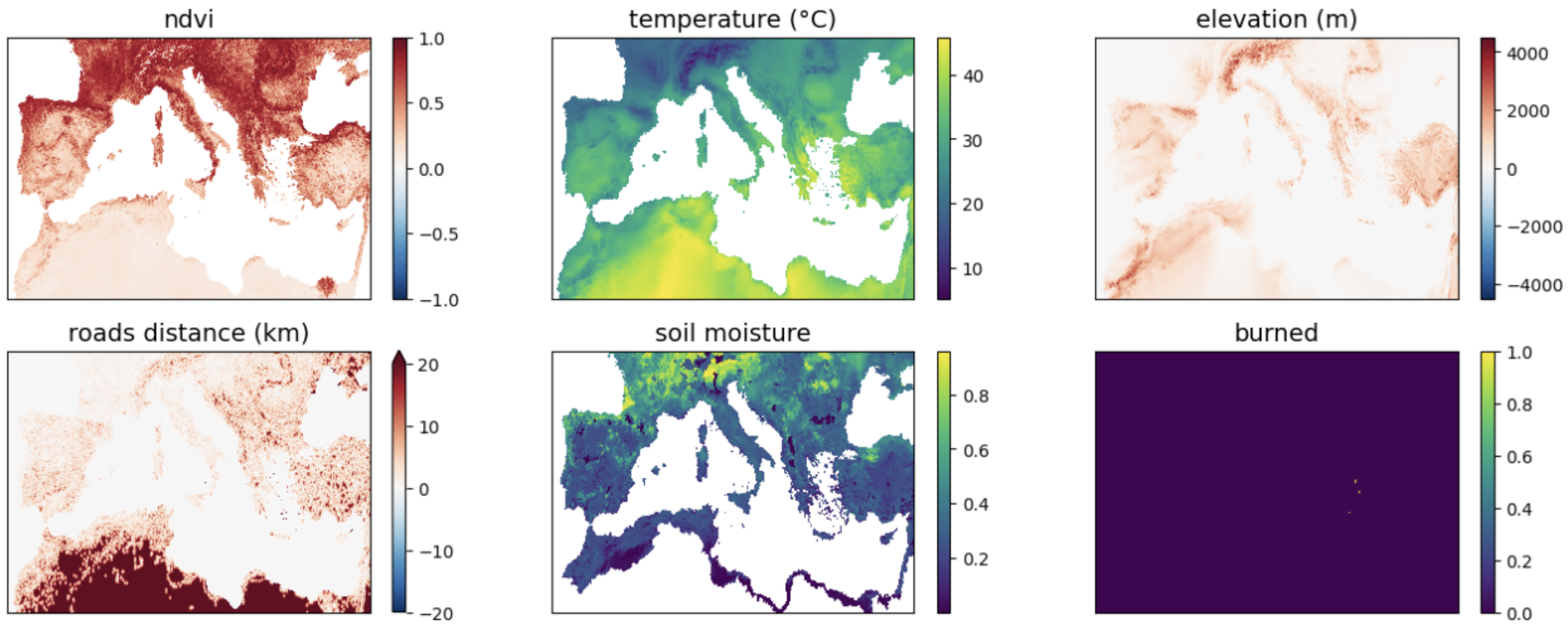}
    \caption{Visualization of some of the variables in Mesogeos datacube for a day.}
    \label{fig:variables}
\end{figure}

Mesogeos was developed by the Orion Lab of the Institute for Astronomy, Astrophysics, Space Applications, and Remote Sensing of the National Observatory of Athens in collaboration with the Image Processing Laboratory (IPL) of the University of Valencia.
This work has received funding from the European Union’s Horizon 2020 Research and Innovation Projects DeepCube and TREEADS, under Grant Agreement Numbers 101004188 and 101036926 respectively.

\subsubsection{Composition}
The Mesogeos dataset is structured in a spatio-temporal grid format, namely a datacube with three dimensions: longitude, latitude, and time. 
The datacube encompasses 27 variables related to known fire drivers such as meteorology, vegetation, land cover, and human activity. 
Mesogeos also includes historical burned areas, ignitions, and burned area sizes as separate variables. 
It has $1km\times1km\times daily$ resolution and contains the values of the variables covering the period from $2006$ to $2022$. 
It incorporates data from the wide Mediterranean area and spans a total area of $4714 km \times1753 km$ and $6026$ days, for a total of $47.796.706.692$ different data points for each dynamic variable.

The detailed list of the variables with their input sources, original temporal and spatial resolution, and their units are provided in Table \ref{tab:variables}.

We present a visualization of some of the variables in the datacube for a day in Figure \ref{fig:variables}.

\begin{table}[htbp]
    \caption{Table of all the variables in Mesogeos.}
    \centering
    \begin{tabular}{lcccc}
        \toprule
        Variable & Source & Sp. Res. & Temp. Res. & Units \\
        \midrule
        \multicolumn{5}{c}{\textit{Dynamic variables}} \\
        \midrule
        Max Temperature & ERA5-Land & 9km & hourly & K \\
        Max Wind Speed & ERA5-Land & 9km & hourly & m/s \\
        Max Wind Direction & ERA5-Land & 9km & hourly & $^{\circ}$ \\
        Max Dewpoint Temperature & ERA5-Land & 9km & hourly & K \\
        Max Surface Pressure & ERA5-Land & 9km & hourly & Pa \\
        Min Relative Humidity & ERA5-Land & 9km & hourly & $\%/100$ \\
        Total Precipitation & ERA5-Land & 9km & hourly & m \\
        Mean Surface Solar Radiation Downwards & ERA5-Land & 9km & hourly & $J/m^2$ \\
        Day Land Surface Temperature & MODIS & 1km & daily & K \\
        Night Land Surface Temperature & MODIS & 1km & daily & K \\
        Normalized Difference Vegetation Index (NDVI) & MODIS & 500m & 16-days & - \\
        Leaf Area Index (LAI) & MODIS & 500m & 8-days & - \\
        Soil moisture & EDO & 5km & 10-days & - \\
        Burned Areas & EFFIS & 1km & vector & $\{0,1\}$ \\
        Ignition Points & MODIS & 1km & vector & hectares \\

        \midrule
        \multicolumn{5}{c}{\textit{Semi-static variables}} \\
        \midrule
        Population & Worldpop & 1km & yearly & $people/km^2$ \\
        Fraction of agriculture & Copernicus CCS & 300m & yearly & $\%/100$ \\
        Fraction of forest & Copernicus CCS & 300m & yearly & $\%/100$ \\
        Fraction of grassland & Copernicus CCS & 300m & yearly & $\%/100$ \\
        Fraction of settlements & Copernicus CCS & 300m & yearly & $\%/100$ \\
        Fraction of shrubland & Copernicus CCS & 300m & yearly & $\%/100$ \\
        Fraction of sparse vegetation & Copernicus CCS & 300m & yearly & $\%/100$ \\
        Fraction of water bodies & Copernicus CCS & 300m & yearly & $\%/100$ \\
        Fraction of wetland & Copernicus CCS & 300m & yearly & $\%/100$ \\
        \midrule
        \multicolumn{5}{c}{\textit{Static variables}} \\
        \midrule
        Roads distance & Worldpop & 1km & static & km \\
        Elevation & COP-DEM & 30m & static & m \\
        Slope & COP-DEM & 30m & static & rad \\
        Aspect & COP-DEM & 30m & static & $^{\circ}$ \\
        Curvature & COP-DEM & 30m & static & rad \\
        \bottomrule
    
    \end{tabular}
    \label{tab:variables}
\end{table}

\begin{figure}
    \centering
    \includegraphics[width=\linewidth]{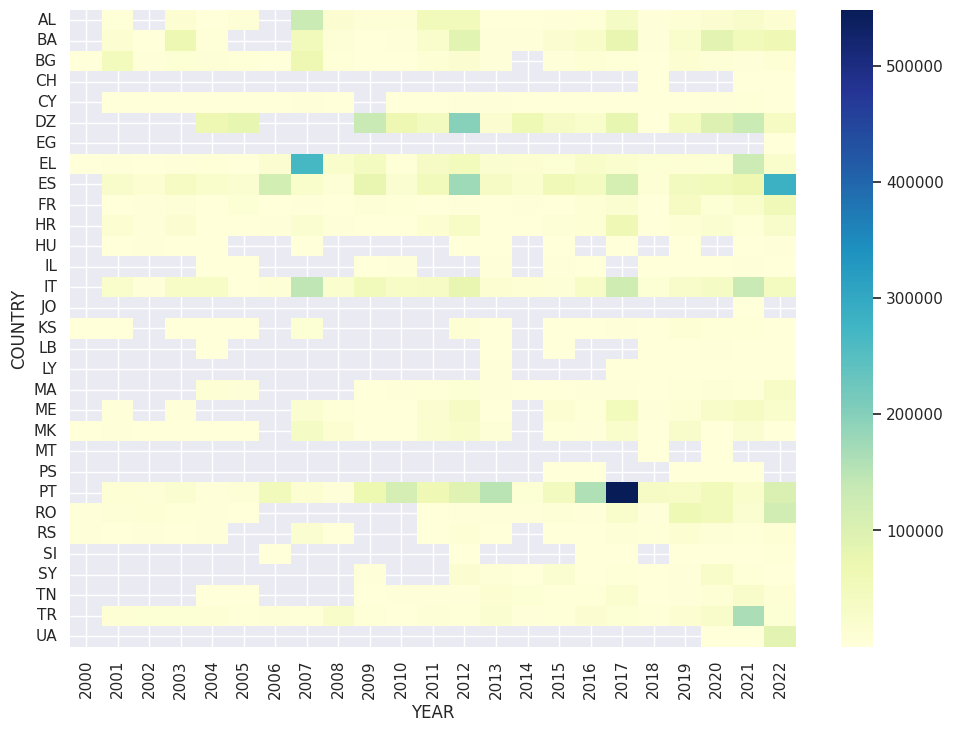}
    \caption{The distribution of the number of fires per country in the burned areas product provided by EFFIS.}
    \label{fig:countries}
\end{figure}

Mesogeos inherits inaccuracies of the original data sources.
Many of the data come from satellites (e.g. burned areas and fire ignitions, land surface temperature (LST), NDVI, LAI. 
Thus satellites’ spatial resolution and missing data from cloud cover cascade to the datacube. 
Moreover, due to satellites' unavailability for specific days, there are some data missing for the LST MODIS product that was replaced by the values of the previous days (172nd of 2006, 190th of 2015, 265th of 2021, 344th of 2019, 85th of 2017, 169th of 2019, the dates are in Julian date format). The same holds for the 233rd day of 2019 for the LAI product.
Moreover, the burned areas dataset provided by EFFIS does not offer the same coverage of burned areas across all Mediterranean countries throughout the specified timeframe, resulting in the absence of data for certain countries in specific years.
% However, the burned areas product from EFFIS is still the most accurate available product for wildfires in Europe.
The distribution in the years from the EFFIS product is provided in Figure \ref{fig:countries}.
However, EFFIS is continuously post-processing fires from the countries spanning in the past, thus when they are made available, our intention is to append them in the datacube.
In the meanwhile, if as more accurate products that refer to fires are produced, they can easily be appended to the datacube as separate variables.

As also noted in the main text according to the EFFIS website, the start dates of the fires provided by EFFIS may not always correspond to the date of ignition of the fire. For this, we make a specific post-processing to acquire the first detection of a wildfire, as this is very important in wildfire modeling. 
Despite the improvements achieved through our approach, it is essential to recognize that some misalignments in the ignition dates may persist in the datacube.

From the multi-purpose dataset Mesogeos, in this work, we extract two ML-ready datasets which are created by sampling the original datasets.
One tailored to fire danger forecasting and one to final burned area prediction.
The sampling scheme for the two datasets is analytically presented in the main paper.
Moreover, the training, validation, test splits are also provided in the main text.

\subsubsection{Collection Process}

All the data in Mesogeos were collected directly from the data sources in the format originally provided by each data source, and a recent crawling was used close to the final date of the datacube, September 29, 2022.
The collection process for each of the data sources used for Mesogeos was different for each product:

\begin{itemize}
    \item \textbf{ERA5-Land:} The hourly data from ERA5-Land were collected from \href{https://cds.climate.copernicus.eu/cdsapp\#!/dataset/reanalysis-era5-land?tab=form}{https://cds.climate.copernicus.eu/cdsapp\#!/dataset/reanalysis-era5-land?tab=form} using the API request tool provided by the platform. Access to data is based on a principle of full, open, and free access (\href{https://cds.climate.copernicus.eu/api/v2/terms/static/licence-to-use-copernicus-products.pdf}{https://cds.climate.copernicus.eu/api/v2/terms/static/licence-to-use-copernicus-products.pdf}). The days and the timespan used was decided to match the shape of the datacube.
    \item \textbf{MODIS:} The data from MODIS (day's/night's land surface temperature, NDVI, LAI) were downloaded from NASA's portal \href{https://modis.gsfc.nasa.gov/data/}{https://modis.gsfc.nasa.gov/data/} using the official python script provided for downloading that needs to declare the days, tiles and products that need to be downloaded. There are no restrictions on the use of data (\href{https://lpdaac.usgs.gov/data/data-citation-and-policies/}{https://lpdaac.usgs.gov/data/data-citation-and-policies/}). We downloaded the data covering the temporal and spatial dimensions of the datacube. Specifically, we downloaded tiles h17v04, h18v04, h19v04, h20v04, h17v05, h18v05, h19v05, and h20v05 for all the years from 2006 to 2022. 
    \item \textbf{European Drought Observatory (EDO):} The soil moisture index was downloaded by a simple wget command from the EDO website \href{https://edo.jrc.ec.europa.eu/gdo/php/index.php?id=2112}{https://edo.jrc.ec.europa.eu/gdo/php/index.php?id=2112}. The data are distributed under a free open access (\href{https://edo.jrc.ec.europa.eu/gdo/php/index.php?id=2044}{https://edo.jrc.ec.europa.eu/gdo/php/index.php?id=2044}). We downloaded the whole data covering the timespan of the datacube. 
    \item \textbf{European Forest Fire Information System (EFFIS):} The EFFIS data were downloaded from the real-time updated burnt areas database of the Data and Services tab of the EFFIS website \href{https://effis.jrc.ec.europa.eu/applications/data-and-services}{https://effis.jrc.ec.europa.eu/applications/data-and-services}. The data are distributed under the Creative Commons Attribution 4.0 International License. From all the data we used the data from 2006 til 2022, as these are the data with corrected dates. The previous years have not been processed yet from the EFFIS, as we noticed that all dates of all the fires were dated as 01/01/year.
    \item \textbf{Ignition Points:} The ignition points were downloaded from NASA's FIRMS download archive \href{https://firms.modaps.eosdis.nasa.gov/download/create.php}{https://firms.modaps.eosdis.nasa.gov/download/create.php} portal by a manual request for the specific region of interest and timespan. There are no restrictions on the use of data (\href{https://lpdaac.usgs.gov/data/data-citation-and-policies/}{https://lpdaac.usgs.gov/data/data-citation-and-policies/}).
    \item \textbf{Land Cover:} The land cover data were downloaded from \href{https://cds.climate.copernicus.eu/cdsapp\#!/dataset/satellite-land-cover?tab=form}{https://cds.climate.copernicus.eu/cdsapp\#!/dataset/satellite-land-cover?tab=form} for the years 2006-2022 using the API request tool provided by the platform. Access to data is based on a principle of full, open and free access \href{https://cds.climate.copernicus.eu/api/v2/terms/static/licence-to-use-copernicus-products.pdf}{https://cds.climate.copernicus.eu/api/v2/terms/static/licence-to-use-copernicus-products.pdf}. 
    \item \textbf{Worldpop data:} Roads distance and population data were downloaded via wget from \href{https://www.worldpop.org}{https://www.worldpop.org} for each country in the datacube's spatial span separately. Then the data from the separate countries were merged in a single grid using the GDAL (\href{https://gdal.org/}{https://gdal.org/}) merge command. The data are distributed under the Creative Commons Attribution 4.0 International License. The population data were provided yearly, thus the data that were downloaded were from 2006 to 2022, while for the road distance product, there is one single instance. 
    \item \textbf{Digital Elevation Models:} The elevation product was downloaded from COP-DEM \href{https://spacedata.copernicus.eu/en/web/guest/collections/copernicus-digital-elevation-model}{https://spacedata.copernicus.eu/en/web/guest/collections/copernicus-digital-elevation-model} using a wget script. The product is available under a free license \href{https://spacedata.copernicus.eu/collections/copernicus-digital-elevation-model}{https://spacedata.copernicus.eu/collections/copernicus-digital-elevation-model}. Then a post-process was used to generate slope, aspect, and curvature as described in the next section. This product is static. 
\end{itemize}

It is unknown to the authors of the dataset if any ethical review processes were conducted by the providers of the datasets.

\subsubsection{Pre-processing, Cleaning, Labeling}

We store the collected data in an internal machine and post-process them in order to append them in the $1$ km $\times$ $1$ km $\times$ $1$ day resolution datacube.

\begin{itemize}
    \item \textbf{ERA5-Land Data.} We combine 2 m dewpoint temperature (DT) and 2 m temperature (T), to calculate relative humidity using the following equation:
    \begin{equation}
        100*\frac{\exp{\frac{17.625*DT}{243.04+DT}}}{\exp{\frac{17.625*T}{243.04+T}}}
    \end{equation}
    We also combine 10 m wind $u$-component and 10 m wind $v$-component to compute wind speed and direction. 
    Then, we calculate the maximum daily value of 2 m temperature, 2 m dewpoint temperature, and surface pressure, the minimum daily value of relative humidity, the mean value of the surface solar radiation downwards, and the day's total precipitation. 
    All these are calculated based on the hourly values of the day. 
    For the wind, we calculate daily values of maximum wind speed.
    The daily wind speed direction is calculated at the hour that the maximum wind speed occurs. 
    Finally, we use linear interpolation to map the 9 km spatial resolution to the 1 km spatial resolution of the datacube.
    \item \textbf{LAI, NDVI.} We concatenate the separate tiles downloaded from MODIS and we use the nearest interpolation to map these variables to 1 km spatial resolution. 
    Moreover, as the time resolution of the products is 8 and 16 days, we forward-fill the values in time, to fill the temporal gaps and let the variables have a daily temporal resolution. 
    \item \textbf{Day and Night Land Surface Temperature.} No pre-processing was conducted for these products.
    \item \textbf{Soil Moisture Index}. We reproject this variable in EPSG:4326 and we use the nearest interpolation to map it to 1 km spatial resolution.
    As the temporal resolution of the variables is 10 days, we also forward-fill it in time, to fill the temporal gaps. 
    \item \textbf{Roads Distance} is a static variable that has no time dimension. No pre-processing is needed. 
    \item \textbf{Yearly population density}. We collect each year's population density covering the years 2006-2020. 
    As the years 2021, and 2022 are not available, we use the year's 2020 value as a proxy for them. 
    \item \textbf{Elevation, Slope, Aspect, Curvature}. After gathering the elevation, we upscale it to a 1 km spatial resolution by using mean aggregation. 
    After having the 1 km spatially resolved elevation at hand, we calculate slope, aspect, and curvature at 1 km spatial resolution, using GDAL \url{https://gdal.org/}.
    \item \textbf{Land Cover}. The land cover product comes in 300 m spatial resolution. 
    In order to take a 1 km spatial resolution product, we create 8 variables, each one related to the fraction of one out of $8$ classes of interest, which are based on the following subclasses: agriculture, forest, grassland, settlements, shrubland, sparse vegetation, water bodies, wetland.
    For this, we calculate the percentage of each subclass in the 1km pixel of interest.
    Thus, in each 1 km spatially resolved cell, each of these $8$ variables has a value between $0$ and $1$, which represents the fraction of the class presence in the pixel.
    We repeat this process for all the years.
    \item \textbf{Burned Areas \& Ignition Points.} The gathered shapefiles representing final burned areas, produced by different ignitions need some pre-processing to identify the date of the ignition of the fire and to combine several burned areas, that are produced by the same fire event, into one. 
    To this end, we associate burned area products with the product of active fires. 
    In order to identify the date of the ignition, we create a 1 km buffer around the burned area shapefiles and we look for an active fire inside this area that has been produced within a week's temporal buffer from the date that EFFIS provides for the burned area. 
    Moreover, to combine several burned areas into one, we also use a buffer of 1 km and we do the following: We use as an anchor a burned area and we calculate its ignition point following the procedure above. 
    If this ignition point is also lying inside buffered burned areas other than the anchor one and these burned areas have a date within a week before the anchor one, we consider that they belong to the same event as the anchor one. 
    Having the ignition points and burned areas at hand, we rasterize them in the same 1 km spatial grid of all the input variables.
    Each ignition point also holds the burned area size of the fire.
\end{itemize}

The final datacube has 29 variables. 

We shift backward the variables related to fire events (burned areas, ignition points). Moreover, we also shift backward the meteorological data, as we assume that will be available through forecasts for the day of interest.

\subsection{Models Details}

\subsubsection{Track A: Next Day's Wildfire Danger Forecasting}

Long Short-term Memory (LSTM), Transformer, and Gated Transformer are the models used for Track A. All the hyperparameters included in this section were defined using the best-performing model in the validation set. All the models were trained for 30 epochs with the binary cross-entropy loss with $\ell_2$-norm regularization, the Adam optimizer, and a batch size of $256$. The data were normalized before serving as input. 
We filled with the temporal aggregate of the time series the nulls existing in the inputs of the models. 
When the value was still null, we were filling it with 0, which is the mean of each variable, after being normalized. 

Regarding LSTM's architecture, a normalization layer is followed by an LSTM layer with $128$ neurons, which is followed by two linear hidden layers with $128$ and $64$ neurons, respectively, and an output $2$-class softmax layer. The final model is trained with a $0.0063$ weight decay and $0.004$ learning rate, which is divided by 10 every 15 epochs. 
All linear layers, but the last, are followed by a dropout with probability $p=0.5$ and the ReLU activation function. 
% The model was trained in an NVIDIA GeForce RTX 3080 GPU on an internal server.

Regarding Transformer's architecture, the time-series input is passed through a linear layer with $256$ neurons and then through a standard positional encoding layer. Then, it is followed by $2$ standard Transformer Encoder Layers which are made up of self-attention and feed-forward layers. The number of heads in each layer is $2$ and the neurons in each feed-forward layer are $512$. 
The final layer is a $2$-class softmax layer. 
The final model is trained with a $0.0018$ weight decay and $0.00029$ learning rate, which is divided by 10 every 15 epochs. 

Regarding Gated Transformer's architecture, for the time-Transformer block, the original time-series input is passed through a linear layer with $256$ neurons and then through a standard positional encoding layer. 
This is followed by $4$ standard Transformer Encoder Layers which are made up of self-attention and feed-forward layers. 
The number of heads in each layer is $4$ and the neurons in each feed-forward layer are $512$. 
Regarding the channel-Transformer block, following the original paper, the time-series input is transposed in order for the attention to act in the dimension of the variables. 
The input is then passed through $4$ Transformer Encoder Layers with the same parameters as the time one.
The outputs of the two Transformers blocks are then concatenated and passed through a softmax layer.
The first logit of the softmax is multiplied with the outputs of time-Transformer and the second by the output of the channel-Transformer.
The final outputs are concatenated and pass through the final $2$-class softmax layer. 
The final model is trained with a $0.0045$ weight decay and $0.00012$ learning rate, which is divided by 10 every 15 epochs. 

All the models were trained in a system with 2 GPUs (NVIDIA GeForce RTX 3080), each one having a memory of 10 GB. The total RAM of the system is 128GB and it also has 48 CPUs.
The models were trained using one of the GPUs.

All models were developed using the PyTorch library \cite{pytorch}

\subsubsection{Track B: Final Burned Area Prediction}

For track B, the U-Net was employed.

A standard U-Net model was used with an EfficientNet-B1 Encoder, using Pytorch \cite{pytorch}, Pytorch Ligthning \cite{falcon2019pytorch} and the segmentations\_model\_pytorch library \cite{Iakubovskii:2019}.
The model was trained for a maximum of $50$ epochs, using early stopping, with the binary cross-entropy loss with $\ell_2$-norm regularization, the Adam optimizer, and a batch size of $128$.
The standard U-Net architecture was used with no modifications.
The final model is trained with a $0.000001$ weight decay and $0.005$ learning rate.
The data are scaled with the minimum and maximum values in the range [0, 1] before being served as inputs to the model.
We fill the nulls with -1.
The model was trained in a system with 2 GPUs (NVIDIA GeForce RTX 3090), each one having a memory of 24 GB. The total RAM of the system is 128GB and it also has 32 CPUs.
The models were trained using one of the GPUs.

\end{document}